\documentclass[conference]{IEEEtran}
\IEEEoverridecommandlockouts
\usepackage{cite}
\usepackage{stfloats}

\usepackage{amsmath,amssymb,amsfonts}
\usepackage{algorithmic}
\usepackage{graphicx}
\usepackage{textcomp}
\usepackage{xcolor}
\def\BibTeX{{\rm B\kern-.05em{\sc i\kern-.025em b}\kern-.08em
    T\kern-.1667em\lower.7ex\hbox{E}\kern-.125emX}}
\begin{document}

\title{A Closed-Loop Multi-Agent System Driven by LLMs for Meal-Level Personalized Nutrition Management
}

\author{
  \IEEEauthorblockN{Muqing Xu*}
  \IEEEauthorblockA{\textit{Dept. Engineering Mathematics}\\University of Bristol\\
  Bristol, UK\\
  hu24149@bristol.ac.uk }
}

\maketitle

\begin{abstract}
Personalized nutrition management aims to tailor dietary guidance to an individual’s intake and phenotype, but most existing systems handle food logging, nutrient analysis and recommendation separately. We present a next-generation mobile nutrition assistant that combines image based meal logging with an LLM driven multi agent controller to provide meal level closed loop support. The system coordinates vision, dialogue and state management agents to estimate nutrients from photos and update a daily intake budget. It then adapts the next meal plan to user preferences and dietary constraints. Experiments with SNAPMe meal images and simulated users show competitive nutrient estimation, personalized menus and efficient task plans. These findings demonstrate the feasibility of multi agent LLM control for personalized nutrition and reveal open challenges in micronutrient estimation from images and in large scale real world studies.
\end{abstract}


\section{Introduction}
 Personalized nutrition is drawing growing attention in healthcare and disease prevention. It can improve health and help prevent and control chronic diseases such as diabetes and obesity. However, traditional dietary assessment methods often have limitations. They usually rely on self-report or interviews with dietitians \cite{b1}, such as 24-hour dietary recall (24HR), diet records (DR), and food frequency questionnaires (FFQ). For most people without nutrition training, the reported food types and portion sizes in these methods depend largely on the user’s own judgment, which can lead to bias and inaccuracy in the results. This is where mobile health (mHealth) can offer a promising solution.

 mHealth is generally defined as healthcare and public health practice supported by mobile devices, wireless sensors, and communication technologies \cite{b2}. In nutrition management, mHealth has greatly improved dietary assessment and interventions. Compared with traditional methods, mobile apps can provide faster and more accurate monitoring through food image recognition, barcode scanning, and links to nutrition databases  \cite{b3} \cite{b4}. In addition, with AI and machine learning, they can give personalized nutrition advice based on individual traits and health needs, opening new paths for chronic disease prevention and health maintenance \cite{b5}.

 In recent years, researchers have begun to explore multi-agent systems (MAS) for nutrition management \cite{b6}. In these systems, different agents work together on tasks such as data collection, food recognition, plan generation, and user interaction  \cite{b7}. This approach offers scalability, efficiency, and robustness, while also delivering personalized services to users. It lays the groundwork for next-generation personalized nutrition systems.

 This paper presents a meal-level personalized nutrition guidance system based on a closed-loop multi-agent design, together with a usable mobile application. The system provides personalized dietary advice from multimodal user inputs and supports dynamic nutrition plan adjustment. We use the Dietary Reference Intakes (DRIs) \cite{b8} \cite{b9} \cite{b10} from the U.S. Department of Agriculture (USDA) as the nutrient intake database. Meal images and their nutrition data come from the Surveying Nutrient Assessment with Photographs of Meals (SNAPMe)  \cite{b11} database, and tests are run on the Android Studio Emulator. In Section II, we review related work that connects to our goals. Section III explains the system architecture for multimodal input processing and design, including model choices, communication among agents, and mobile app development. Section IV describes the experimental setup and shows data analysis to evaluate system performance. Finally, Section V concludes the paper.

\section{Related Work}

In recent years, rapid progress in artificial intelligence (AI) has offered promising solutions to long-standing challenges in nutrition management. Among these, vision-based dietary assessment has become a widely used approach. This progress is largely due to the creation and curation of large-scale meal datasets, which serve as key training resources for AI-driven assessment systems. Well-known examples include the Food2K dataset \cite{b12}, the Recipe1M+ dataset \cite{b13}, and the Hd-epic dataset \cite{b14}. Building on rich datasets, many AI algorithms and tools for vision-based dietary assessment have been developed—for example, automatic food recognition\cite{b15} and ingredient recognition \cite{b16}, food segmentation  \cite{b17}, food volume estimation \cite{b18}, recipe retrieval \cite{b19} or generation \cite{b20}, and the use of machine-learning smart sensors to detect eating events \cite{b21} and behaviors \cite{b22} in free-living settings—to provide accurate and convenient daily dietary assessment for everyone. These technologies have greatly changed how dietary intake is evaluated in nutrition management, making many parts of the nutrition ecosystem more efficient and automated, and also shaping how people manage their daily intake \cite{b23}.

Large language models (LLMs) are also playing a growing role in nutrition management. Beyond serving as general nutrition assistants, a paper titled “Dietary Assessment With Multimodal ChatGPT: A Systematic Analysis” \cite{b24} studied the potential of multimodal large language models (MLLMs), especially GPT-4V, for dietary assessment. The study used wearable cameras to collect participants’ intake data and performed food recognition, portion estimation, and nutrition analysis, then combined these results to identify nutrition gaps and recommend foods. The findings show that GPT-4V performs well on these tasks, supporting the feasibility of using MLLMs in nutrition management.

However, many models and algorithms used for dietary assessment still have a limited scope. They often handle only single, discrete tasks, such as recognizing food items or retrieving recipes for a specific dish. These methods lack a full understanding of dietary knowledge, routine practice, assessment workflows, and differences in individual eating behaviors and cultural contexts. As a result, they fall short of the overall goal of accurately quantifying a person’s food and nutrient intake, and they cannot provide truly personalized guidance for healthy eating and health management.

Although multi-agent–based recommender systems have appeared in recent years \cite{b6}\cite{b25}, they do not provide a specialized pipeline and dynamic adjustment strategy for personalized nutrition management. To address this gap, this paper proposes a multi-agent workflow of “meal nutrition analysis – dynamic nutrition plan adjustment – personalized dietary recommendation.”

\section{Closed-loop System Architecture}
\subsection{Overall System Workflow}
\begin{figure}[htbp]
  \centering
  \includegraphics[width=0.8\linewidth]{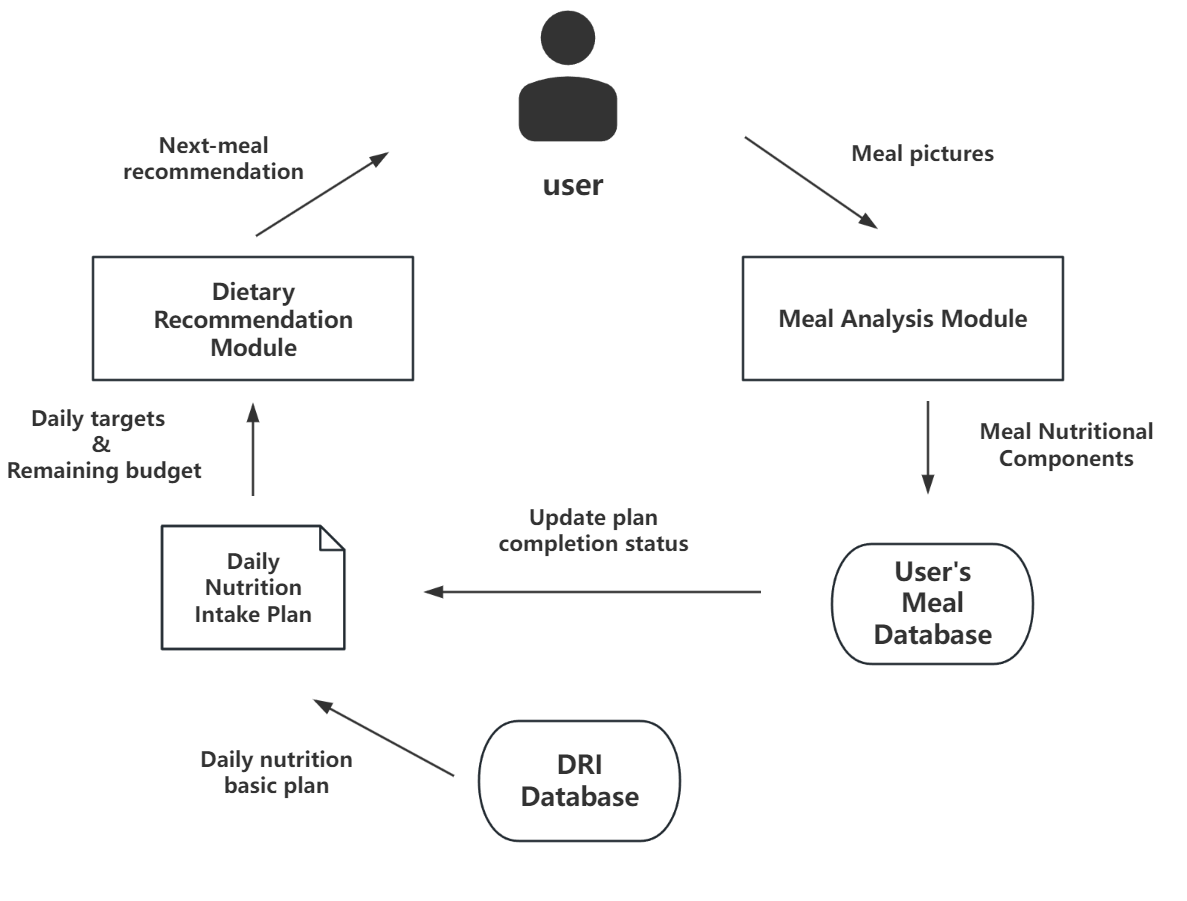}
  \caption{Overview of the system workflow.}
  \label{overall}
\end{figure}

The overall workflow of the system is shown in Fig.~\ref{overall}. Each day starts with a “Daily Nutrition Intake Plan.” Based on the user profile and the Dietary Reference Intakes (DRIs) database, the system generates this plan, producing a set of daily targets and a “remaining allowance” status. When the user sends a meal photo with a short description, the “Meal Analysis Module” converts it into structured nutrition data and writes a time-stamped record to the “User Meal Database.” After each write, the system updates the plan status to reflect what has been consumed and what remains for the day. Then, the “Dietary Recommendation Module” reads the current plan and the latest intake summary and returns a recommendation for the next meal. If no valid image is available or confidence is low, the system uses text-only estimation or asks the user for clarification. This loop continues for every meal during the day and resets at the start of the next day. By separating analysis, state management, and recommendation, the system enforces clear data contracts and supports auditing; the closed loop also limits error propagation, keeping each recommendation aligned with the most recent intake.

\subsection{Closed-Loop Dynamic Adjustment Strategy}
To better support personalization and dynamic adaptation, the recommendation module combines per-meal adjustment with per-day adjustment in the intake plan. In our proposed strategy for dietary recommendation and plan updates, guidance for the next meal is driven by the current plan state, rather than a fixed daily menu. The system takes meal evidence from images and text, maintains an accurate record of actual intake, and then produces actionable advice for the next meal. After each meal, the system aggregates what was actually consumed and updates the day’s remaining nutrition “budget”. Before generating a recommendation, the service reads the latest completion status and allocates the remaining needs across the meals left in the day. Then, using long-term logs and user habit records for preferences and constraints, the system recommends foods and portions that best fill the main nutrient gaps. This design narrows the gap between the “remaining daily targets” and the “consumed amounts” step by step, offering more precise and more personalized nutrition management.

When data are sparse, the recommendation service still provides conservative suggestions. This design treats each meal as a small step toward the goal. Compared with issuing a large, fixed plan in the morning, it reduces user burden and improves adherence.

\begin{figure}[htbp]
  \centering
  \includegraphics[width=0.8\linewidth]{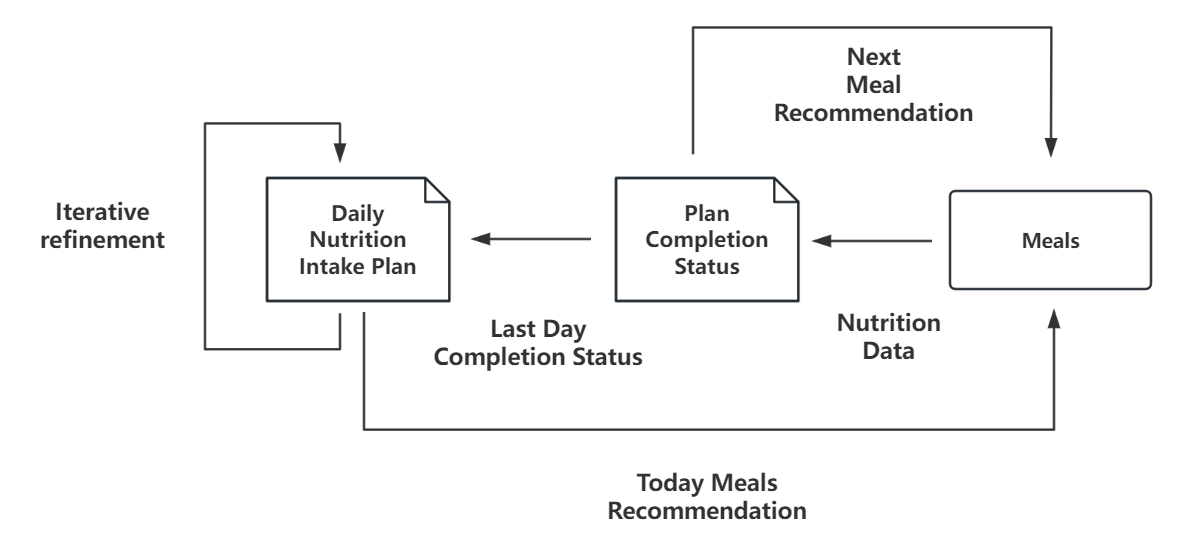}
  \caption{Closed-Loop Dynamic Adjustment Strategy}
  \label{closeloop}
\end{figure}

In sum, we model the task as a closed-loop decision process: each meal updates the system state and influences the next meal’s recommendation. In addition, when a new plan is generated each day, the plan creation service looks at the previous day’s completion and adjusts the new plan accordingly. The meal-level and day-level closed-loop adjustment flow is shown in Fig.~\ref{closeloop}.

\subsection{MAS Architecture}

We decompose the main workflow into several subtasks—meal image recognition, nutrition analysis, and the planning and adjustment of the nutrition plan—and use a MAS so that different agents handle one or more tasks. The MAS proposed in this paper consists of four agents with clear roles: a Controller Agent, a Dialog Agent, a Vision Agent, and a File Agent. The responsibilities and model choices for these four agents are shown in Table ~\ref{tab:agents}. The model selection experiments for the Vision Agent are presented in Section IV.

\begin{table*}[t]
\caption{Agents' Main Duties and LLM Selection}\label{tab:agents}
\begin{center}
\begin{tabular}{|c|c|c|c|}
\hline
\textbf{Agent Name} & \textbf{Main Duties} & \textbf{LLM Selection} & \textbf{Rationale} \\
\hline
Controller agent &
\begin{tabular}[c]{@{}c@{}}Planner and coordinator; categorizes requests\\ and selects the shortest, safest workflow.\end{tabular} &
GPT-4o-mini &
\begin{tabular}[c]{@{}c@{}}Excellent reasoning and multi-step planning\\ while balancing latency and cost.\end{tabular} \\
\hline
Vision agent &
\begin{tabular}[c]{@{}c@{}}Meal picture and nutrient analyzer; converts\\ meal pictures into structured nutrition data.\end{tabular} &
GPT-4o &
\begin{tabular}[c]{@{}c@{}}Strong performance in food recognition and\\ mass/volume estimation; robust to occlusion.\end{tabular} \\
\hline
File agent &
\begin{tabular}[c]{@{}c@{}}Sole writer and administrator of the system state;\\ updates files with conversation content.\end{tabular} &
Claude Sonnet 4 &
\begin{tabular}[c]{@{}c@{}}Stronger instruction following, better formatting\\ compliance, stable tool invocation, lower hallucination.\end{tabular} \\
\hline
Dialog agent &
\begin{tabular}[c]{@{}c@{}}Nutrition consultant; generates next-meal\\ recommendations based on the plan.\end{tabular} &
GPT-4.1-mini &
\begin{tabular}[c]{@{}c@{}}Fast and cost-efficient; sufficient for reading small\\ data files and simple allocation/analysis.\end{tabular} \\
\hline
\end{tabular}
\end{center}
\end{table*}

\begin{figure}[htbp]
  \centering
  \includegraphics[width=0.9\linewidth]{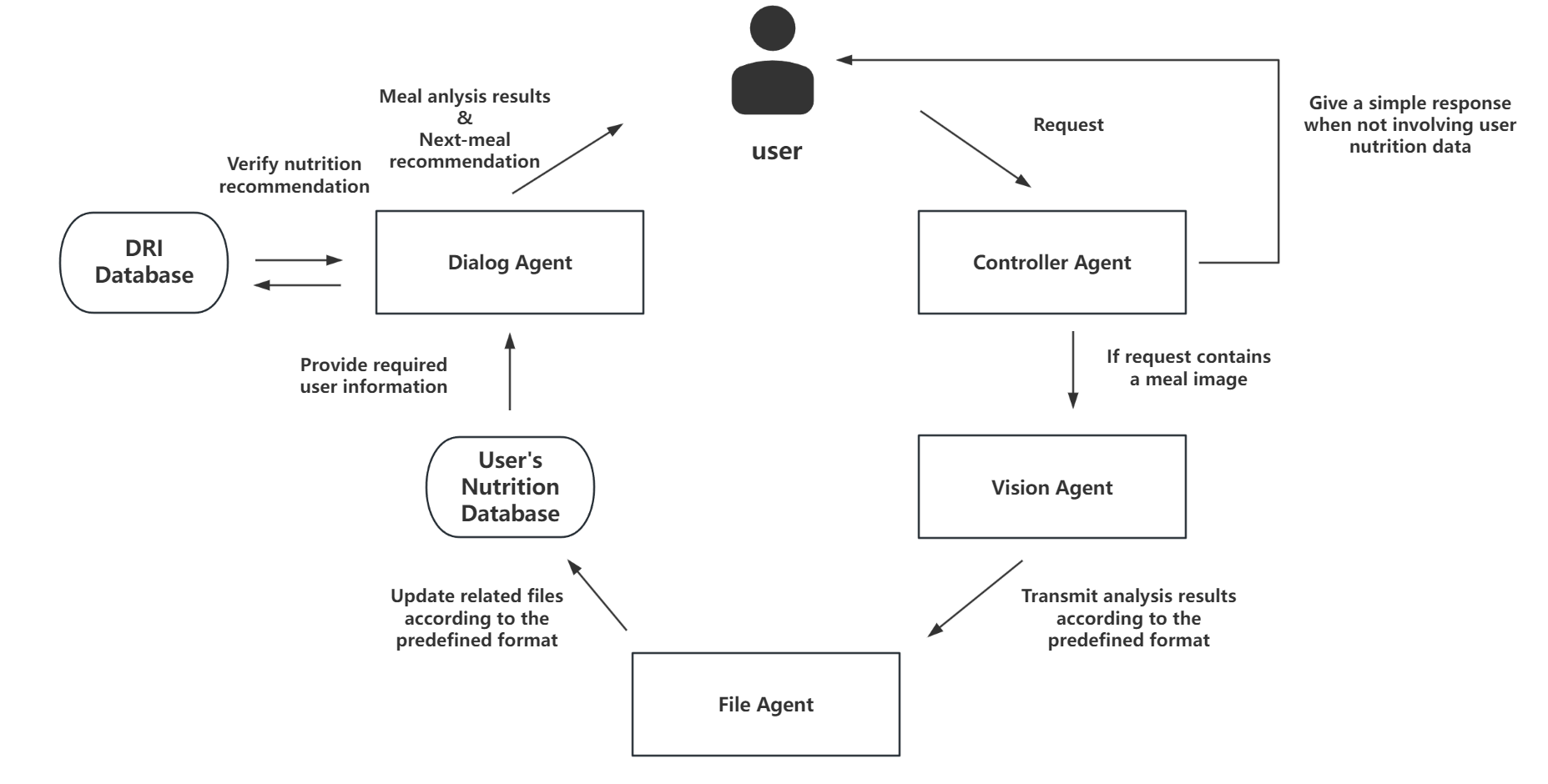}
  \caption{Overview of the MAS Multimodal Main Workflow}
  \label{MAS}
\end{figure}

The MAS exposes a single entry point for user interaction through the Controller Agent. Each user request is packed into a message that includes the user ID, date, mealtime, optional image and text, and a meal ID. From the outset, we ask the system to answer both special nutrition questions and general questions. To avoid wasting tokens, the Controller classifies each text at intake. If the query is general or a light nutrition question that does not involve personal data, the Controller answers directly. If the information concerns nutrition management, the Controller breaks the task down, selects the shortest safe workflow, and schedules the steps. It then delegates each step to the other agents. Communication among agents uses a predefined payload format, so interfaces remain stable and easy to audit. An overview of the MAS multimodal main workflow is shown in Fig.~\ref{MAS}.

\section{Experiments and Results Analysis}

\subsection{Datasets}
Dietary Reference Intakes (DRIs) are science-based reference values set by the Food and Nutrition Board of the U.S. National Academies to support diet planning and assessment for healthy people. These tables are published and compiled for practitioners through the USDA National Agricultural Library (NAL). The DRIs include detailed daily intake tables covering 40 essential nutrients and elements, grouped into three categories: essential minerals, vitamins, and macronutrients. The DRI framework provides several numbers for different purposes: the Estimated Average Requirement (EAR) is used to assess whether a group’s intake is adequate; the Recommended Dietary Allowance (RDA) is used for individual targets. For energy, the DRIs provide the Estimated Energy Requirement and the Acceptable Macronutrient Distribution Range.

In this paper, we selected the RDA tables for essential minerals, vitamins, and macronutrients from the USDA databases. These tables were merged into a single data file as the basis for planning daily nutrient intake. The RDA values are stratified by life stage and sex to reflect physiological and needs differences. Before merging, we cleaned the numeric fields in each table using consistent rules and adjusted units for some nutrients. This process produced three clean tables that share the same primary key and naming scheme. Next, we performed a full outer join across the Category and life stage dimensions of the three tables. After merging, each life stage is represented by one row containing all nutrients side by side. In the system workflow, this file serves as a reliable external reference when generating daily intake targets.

SNAPMe database is a benchmark dataset for photo-based dietary assessment. The study collected meal photos in free-living settings and linked them to concurrent 24-hour food records, allowing model outputs to be compared with structured self-reports. The dataset contains 3,311 unique food photos from 95 participants, covering 275 eating days. In our experiments, we used SNAPMe to compare different large language models on meal image analysis, and we evaluated their abilities mainly by the accuracy of food recognition, volume and mass estimation, and nutrient analysis.

\subsection{Meal Image Analysis Capability}
Given that the accuracy of meal image analysis is a key part of personalized nutrition management, we ran a comparative evaluation of several mainstream multimodal large language models before building the Vision Agent. We tested their performance on the SNAPMe dataset and on a labeled local sample of meal photos, focusing on their ability in food category recognition and volume/mass estimation to make the final model choice. For the food recognition test, we used 21 images covering 50 food items; for the volume/mass estimation test, we used 30 images with a reference object and 30 images without a reference object. The results are shown in Table  ~\ref{tab:food-rec} and Table ~\ref{tab:error-ranges}.

\begin{table}[htbp]
\caption{Comparison of Food Recognition Capabilities Across LLMs}
\begin{center}
\setlength{\tabcolsep}{3pt}
\renewcommand{\arraystretch}{1.05}
\begin{tabular}{|l|c|c|c|}
\hline
\textbf{Model} &
\shortstack{\textbf{Food}\\ \textbf{recognition}\\ \textbf{accuracy}} &
\shortstack{\textbf{Max}\\ \textbf{occlusion}\\ \textbf{rate}$^{\mathrm{a}}$} &
\shortstack{\textbf{Food}\\ \textbf{composition}\\ \textbf{accuracy}} \\
\hline
GPT\mbox{-}4o           & 88\% & $>80\%$ & $\sim$75\% \\ \hline
Qwen QVQ\mbox{-}MAX     & 72\% & 60\%    & $\sim$80\% \\ \hline
Gemini 2.5 Flash        & 92\% & $>80\%$ & $\sim$80\% \\ \hline
Claude Sonnet 4         & 64\% & 60\%    & $\sim$70\% \\ \hline
\multicolumn{4}{l}{$^{\mathrm{a}}$ It represents the highest level of occlusion for food items.}
\end{tabular}
\label{tab:food-rec}
\end{center}
\end{table}

GPT-4o and Gemini 2.5 Flash show clearly stronger overall performance than Qwen QVQ-MAX and Claude Sonnet 4 on food recognition. Both maintain high accuracy and remain reliable under heavy occlusion (more than 80\%). Qwen QVQ-MAX has a slight advantage on East Asian dishes, but its robustness to occluded items is weak and caps at about 60\%. Claude Sonnet 4 is conservative in detection. It often reports only the dishes’ primary ingredient. This behavior hurts recall on composite or complex dishes. Therefore, only GPT-4o and Gemini 2.5 Flash were tested in the subsequent volume/mass estimation test.

\begin{table}[htbp]
\caption{Error ranges and means for food-related estimations}
\begin{center}
\setlength{\tabcolsep}{3pt}
\renewcommand{\arraystretch}{1.05}
\footnotesize
\begin{tabular}{|l|c|c|c|c|}
\hline
\textbf{Category} &
\multicolumn{2}{c|}{\textbf{GPT-4o}} &
\multicolumn{2}{c|}{\textbf{Gemini 2.5 Flash}} \\
\cline{2-5}
 & \textbf{Range} & \textbf{Mean} & \textbf{Range} & \textbf{Mean} \\
\hline
Volume estimation $^{\mathrm{a}}$        & 3.5\%--12\%    & 7.6\%  & 4.2\%--11.7\%  & 7.3\%  \\
\hline
Mass estimation $^{\mathrm{a}}$          & 11.2\%--24.5\% & 17.4\% & 13.3\%--21.9\% & 16.7\% \\
\hline
\shortstack[l]{Total energy\\ estimation $^{\mathrm{a}}$} & 12.6\%--33.6\% & 18.7\% & 13.6\%--30.8\% & 18.2\% \\
\hline
Volume estimation $^{\mathrm{b}}$        & 4.3\%--27.5\%  & 10.7\% & 6.1\%--17.5\%  & 13.8\% \\
\hline
Mass estimation $^{\mathrm{b}}$          & 15.2\%--42.4\% & 26.7\% & 14.7\%--39.9\% & 24.1\% \\
\hline
\shortstack[l]{Total energy\\ estimation $^{\mathrm{b}}$} & 16.4\%--49.7\% & 28.9\% & 18.5\%--45.8\% & 27.4\% \\
\hline
\multicolumn{5}{l}{$^{\mathrm{a}}$ with reference object;\quad $^{\mathrm{b}}$ without reference.} \\
\end{tabular}
\label{tab:error-ranges}
\end{center}
\end{table}
Based on the test results, Gemini 2.5 Flash was slightly better than GPT-4o in estimating food mass, volume, and energy. However, because Gemini occasionally refused to perform volume and mass estimation during testing, we finally chose GPT as the model driving the Vision Agent to ensure workflow stability. Both models also performed noticeably better on tests with a reference object than on tests without one. Therefore, the system prompt for the Vision Agent was written to stress the importance of using a reference object: it instructs the agent to either focus on any reference mentioned in the user prompt before analysis or try to find common references in the image.

\subsection{Nutritional Analysis Capability}
We consider meals \(i=1,\dots,N\) and nutrients \(j\in\mathcal{J}\).
Let $y_{i,j}$ be the ground-truth amount of nutrient $j$ in meal $i$, and $\hat{y}_{i,j}$ be the predicted amount.

Define
\begin{equation}
N_{j}=\sum_{i=1}^{N} m_{i,j}.
\end{equation}

For each nutrient \(j\), MAE is computed over non-missing predictions:
\begin{equation}
\mathrm{MAE}_{j}
=\frac{1}{N_{j}}
\sum_{i=1}^{N} m_{i,j}\,\bigl|\hat y_{i,j}-y_{i,j}\bigr|.
\end{equation}

Coverage quantifies the proportion of meals with a non-missing prediction:
\begin{equation}
\mathrm{Coverage}_{j}
=\frac{1}{N}\sum_{i=1}^{N} m_{i,j}.
\end{equation}

In this test, we selected 40 non-packaged food images from the SNAPMe dataset to evaluate the Vision Agent’s ability in nutrient analysis. The nutrients were divided into three groups: 
\begin{itemize}
\item Full 40 Fields. This set contains every nutrient tracked by the nutrition plan.
    \item Subset\-1:Core nutritions. This subset focuses on variables that directly drive meal planning and feedback: Total Energy, Protein, Carbohydrate, Fat, Total Fiber, and Sodium.
    \item Subset\-2:Micronutrients. This subset covers vitamins and minerals beyond the core set.
\end{itemize}
The results are shown in Table ~\ref{tab:nutrition}.From the test results, the agent shows a deliberate conservative bias during nutrient analysis, and micronutrients tend to increase the error of nutrient content estimates.

\begin{table}[htbp]
\caption{Nutritional Analysis Capability Test Results}
\begin{center}
\setlength{\tabcolsep}{3pt}
\renewcommand{\arraystretch}{1.05}
\footnotesize
\begin{tabular}{|l|c|c|c|c|c|}
\hline
\textbf{Field Set} & \textbf{MAE (kcal)} & \textbf{MAE (g)} & \textbf{MAE (mg)} & \textbf{MAE (mcg)} & \textbf{Coverage} \\
\hline
Full 40 fields & 65.2 & 7.4 & 332 & 208 & 0.76 \\ \hline
Subset1       & 58.9 & 6.8 & 225 & ---$^{\mathrm{a}}$ & 0.96 \\ \hline
Subset2       & ---$^{\mathrm{a}}$ & ---$^{\mathrm{a}}$ & 318 & 189 & 0.61 \\ \hline
\multicolumn{6}{l}{$^{\mathrm{a}}$ ``---'' denotes not applicable.} \\
\end{tabular}
\label{tab:nutrition}
\end{center}
\end{table}

\subsection{Personalization Capabilities}
\begin{figure}[htbp]
  \centering
  \includegraphics[width=0.8\linewidth]{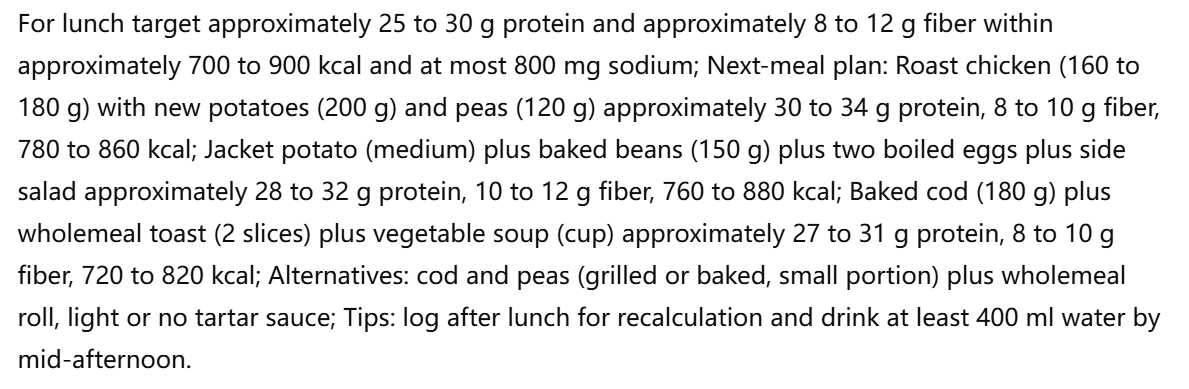}
  \caption{Personalized Next-Meal Recommendations for British Cuisine Preferences}
  \label{bri}
\end{figure}

\begin{figure}[htbp]
  \centering
  \includegraphics[width=0.8\linewidth]{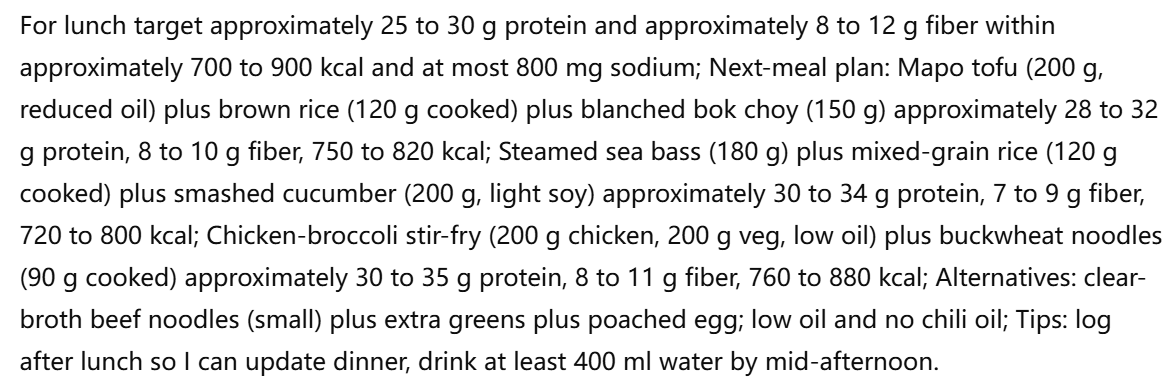}
  \caption{Personalized Next-Meal Recommendations for Chinese Cuisine Preferences}
  \label{chi}
\end{figure}

\begin{figure}[htbp]
  \centering
  \includegraphics[width=0.8\linewidth]{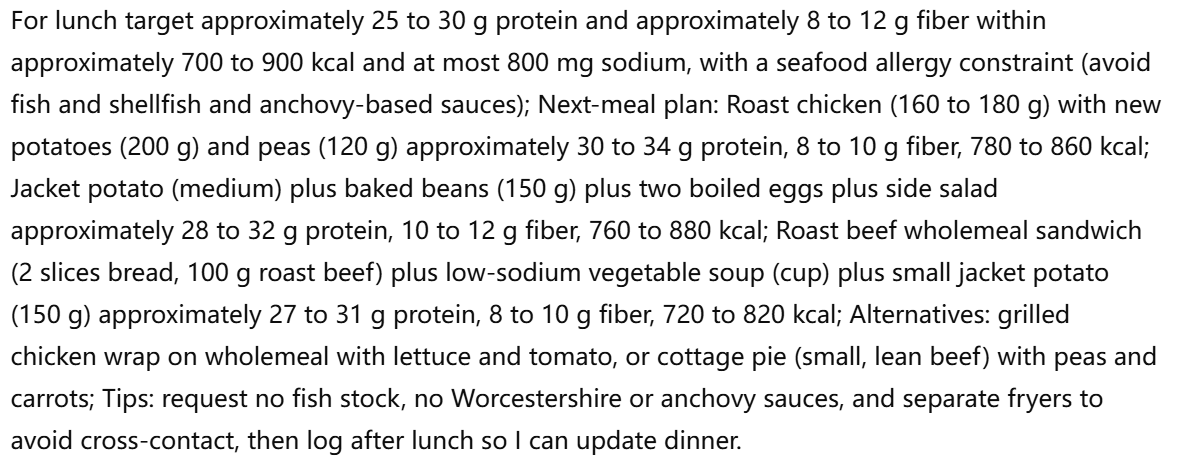}
  \caption{Personalized Next-Meal Recommendations for a user allergic to seafood.}
  \label{bri2}
\end{figure}

In the personalization experiment, we changed the user profile (meal habits and food frequency) by editing the record files, and tested whether the Dialog Agent would adjust its recommendations. With the daily nutrition plan kept the same, we prepared two types of record files—one biased toward Chinese dishes in frequency, and the other biased toward British dishes. Fig. ~\ref{bri} and Fig. ~\ref{chi} demonstrate the comparison of the recommendation results. The outputs indicate that, while meeting the nutrition goals, the dish choices vary with the preference data: one uses Chinese staples, and the other uses British staples. The result shows that the Dialog Agent can use the user profile in the record files to achieve effective personalization without losing goal consistency; the final menus stay aligned in nutrition, differ in style, and remain preference-aware. In the next test, we added “seafood allergy” to the allergy field of the record file and used the British-biased record file. This recommendation from the Dialog Agent is shown in Fig. ~\ref{bri2}.

\subsection{MAS Task Planning and Completion Capability}
In this test, we use Plan Optimality (PO) to evaluate task planning and execution capabilities. This metric is calculated from an executable trace of the agent's tool calls for each task instance. For a task instance \(t\), let \(s(t)\) be the executed step count and \(s^{*}(t)\) be the policy-defined minimal step count.
\begin{equation}
\label{eq:po}
\mathrm{PO}(t)=\frac{s^{*}(t)}{s(t)}, \qquad \mathrm{PO}(t)\in(0,1] .
\end{equation}
A higher \(\mathrm{PO}\) indicates a plan closer to the shortest valid workflow.
We measure latency from server-side timestamps recorded by the Flask API. For a task instance $t$, let $\tau_{\text{in}}(t)$ be the time the request is received and $\tau_{\text{out}}(t)$ be the time the final answer is sent.

\begin{equation}
\label{eq:e2e}
L_{\mathrm{E2E}}(t) \;=\; \tau_{\text{out}}(t) - \tau_{\text{in}}(t) \;.
\end{equation}
In this test, we used 50 multimodal requests with varying complexity, each labeled with its estimated minimum number of steps. Detailed results are shown in Table ~\ref{tab:5.3}. The tests show that the MAS meets its functional goal: returning structured, actionable answers in about 65 s while approaching the shortest valid workflow. Compared with the annotated minimum steps \(s^{*}\) , this means the Controller usually adds about one extra step for every four optimal steps; in the worst case, it is about 1.67 times the shortest path. For comparison, we use a related study as a baseline. First, the navigation literature suggests using the Success weighted by Path Length (SPL) as an efficiency metric equivalent to PO. In Anderson et al.\cite{b26}, reaching 0.5 in unknown, complex environments is considered a reasonable target. By that standard, our PO is higher(0.75).

\begin{table}[t]
\caption{Overall MAS Test Results}
\begin{center}
\setlength{\tabcolsep}{3pt}
\renewcommand{\arraystretch}{1.05}
\begin{tabular}{|l|c|c|c|}
\hline
\textbf{Metric} & \textbf{Average} & \textbf{Maximum} & \textbf{Minimum} \\
\hline
$L_{\mathrm{E2E}}$ & 65.4\,s & 87.8\,s & 53.1\,s \\ \hline
PO                 & 0.75    & 0.86    & 0.60    \\ \hline
\end{tabular}
\label{tab:5.3}
\end{center}
\end{table}

We quantify whether next-day targets are updated in the intended direction given recent unmet residuals.

Let $\mathcal{N}$ be tracked nutrients and $\mathcal{T}=\{1,\ldots,T\}$ days. For $n\in\mathcal{N}$ and $t\in\mathcal{T}$, let $T_n(t)$ and $A_n(t)$ denote the planned target and achieved intake. Define
\begin{equation}
E_n^{(k)}(t)=\frac{1}{k}\sum_{i=0}^{k-1}\big(T_n(t-i)-A_n(t-i)\big).
\end{equation}

\begin{equation}
\Delta T_n(t{+}1)=T_n(t{+}1)-T_n(t).
\end{equation}
and let $s\in\{+1,-1\}$ encode the policy direction. We use Directional Agreement (DA):
\begin{equation}
\Omega=\{(n,t)\mid n\in\mathcal{N},\, t\in\{1,\ldots,T-1\}\},
\end{equation}

\begin{equation}
\mathrm{DA}=\frac{1}{|\Omega|}\sum_{(n,t)\in\Omega}
\mathbf{1}\!\left[\operatorname{sign}(\Delta T_n(t{+}1))=\operatorname{sign}\!\big(sE_n^{(k)}(t)\big)\right]
\end{equation}

We only count pairs with $E_n^{(k)}(t)\neq 0$. We evaluate on 20 synthetic completion scenarios (insufficient less than $-50\%$, near-target between $-5\%$ and $5\%$, excessive greater than $50\%$, with approximately $10\%$ per-nutrient perturbations), reporting mean DA and 95\% CIs.

\begin{table}[htbp]
\caption{Personalized Daily Plan Update Test Results}
\begin{center}
\setlength{\tabcolsep}{3pt}
\renewcommand{\arraystretch}{1.05}
\footnotesize
\begin{tabular}{|l|c|c|}
\hline
\textbf{Method} & $\boldsymbol{\overline{\mathrm{DA}}}$ & \textbf{95\% CI} \\
\hline
MAS             & 0.82 & [0.78, 0.86] \\ \hline
Random baseline & 0.50 & [0.43, 0.57] \\ \hline
Static          & 0.00 & [0.00, 0.00] \\ \hline
\end{tabular}
\label{tab:5.6}
\end{center}
\end{table}

Table~\ref{tab:5.6} shows MAS substantially outperforms both baselines, indicating reliable directional adaptation; DA evaluates direction only.

\section{Conclusion}
This paper proposes an LLM-driven multi-agent system architecture combined with a closed-loop dynamic update strategy to overcome several obstacles in traditional nutrition management. The experimental results show that the proposed system and strategy can effectively provide personalized nutrition guidance. Our study examines the system’s task planning and execution, meal image analysis, and personalization capabilities. We specifically evaluate the agents’ autonomous workflow planning ability, use plan optimality to measure planning efficiency, and investigate the completeness and accuracy of nutrition analysis. However, due to time and cost limits, the evaluation scale is relatively small; in addition, estimating micronutrients from images alone remains highly challenging. Future work will focus on expanding the scope and realism of testing, running controlled user studies to quantify effects on user burden, adherence, and satisfaction. We will also combine image reasoning with lightweight retrieval to improve micronutrient coverage without overstating certainty, bringing the system closer to professional dietitian performance. Recent studies have shown that leveraging LLMs to reason over structured knowledge graphs via SPARQL-style querying can improve efficiency and controllability of knowledge-grounded reasoning, which is a promising direction for reliable nutrition knowledge retrieval in our pipeline\cite{b27}.

\end{document}